\documentclass[conference, a4paper]{IEEEtran}
\IEEEoverridecommandlockouts

\usepackage[english]{babel}
\usepackage[utf8]{inputenc}
\usepackage[T1]{fontenc}
\usepackage{cite}
\usepackage[cmex10]{amsmath}
\usepackage{multirow}
\usepackage{array}
\usepackage[caption=false,font=footnotesize,lofdepth,lotdepth]{subfig}
\usepackage{graphicx}
\usepackage{setspace}
\usepackage{balance}
\usepackage{placeins}
\usepackage{algorithm}
\usepackage{algpseudocode}

\AtBeginDocument{%
  \renewcommand\tablename{TABLE}
  \renewcommand{\thetable}{\arabic{table}}
}

\begin{document}

\title{Hardware-Aware Federated Learning for Speech Emotion Recognition}

\author{%
\IEEEauthorblockN{Beyazıt Bestami YÜKSEL}%
\IEEEauthorblockA{Computer Engineering\\%
Istanbul Technical University\\%
Istanbul, Turkey\\%
yukselbe18@itu.edu.tr}%
\and
\IEEEauthorblockN{Emrah DİKBIYIK}%
\IEEEauthorblockA{Department of Computer Technologies\\%
Istanbul University-Cerrahpaşa\\%
Istanbul, Turkey\\%
emrahdikbiyik@iuc.edu.tr}%
}

\maketitle

\IEEEpubid{\makebox[\columnwidth]{\textbf{979-8-3195-1046-4/26/\$31.00 \copyright2026 IEEE}\hfill}%
\hspace{\columnsep}\makebox[\columnwidth]{}}

\begin{abstract}
Federated learning (FL) enables privacy-preserving collaborative training across distributed edge devices, but real deployments involve heterogeneous clients with different processing power, memory capacity, and communication latency, which often increase round duration and system cost. This paper proposes a hardware-aware federated learning framework for emotion recognition on session-partitioned IEMOCAP that integrates hardware profiling, top-$K$ client selection, and adaptive local epochs within a unified training loop. We compare the method against FedAvg, FedProx, and random top-$K$ selection under a non-IID setup and show that, across 50 federated rounds and 5 independent trials, the proposed approach achieves competitive validation accuracy ($0.352{\pm}0.019$), reduces total training time by about 36.5\% compared to FedAvg, and lowers cumulative communication cost by 40\%.
\end{abstract}

\begin{IEEEkeywords}
Federated learning, hardware-aware scheduling, edge artificial intelligence, speech emotion recognition, distributed machine learning, heterogeneous edge devices.
\end{IEEEkeywords}

\IEEEpubidadjcol

\section{Introduction}

Federated learning is an important paradigm that enables distributed training while keeping data on device and sharing only model updates. It offers a strong alternative for privacy-sensitive domains such as speech analysis and mobile artificial intelligence, as it avoids central collection of raw data.

However, most existing FL algorithms assume that client environments are hardware-homogeneous. In real edge deployments, devices differ substantially in processing power, memory, and network latency; this mismatch causes inefficiency, communication overhead, and the \textit{straggler} effect during training.

This study proposes a hardware-aware FL framework for speech emotion recognition (SER) in heterogeneous edge environments. The main contributions are: (i) a joint approach that unifies client selection, workload adaptation, and latency awareness within a single optimization framework, (ii) a dynamic profiling mechanism that embeds CPU, RAM, latency, and training time into the FL loop for online decision making, (iii) a multi-objective client scheduling strategy under the system-level objective defined in Eq.~(3), (iv) an adaptive workload scaling mechanism that enables stronger devices to contribute proportionally, and (v) empirical validation on a session-based non-IID setup showing improved accuracy together with approximately 36.5\% training-time reduction and 40\% communication-cost reduction.

\section{Related Work}

Federated learning has a broad literature ranging from the foundational FedAvg line \cite{mcmahan2017} to system design, security \cite{YUKSEL2026109557}, and scalability discussions \cite{bonawitz2019,kairouz2021}.

On computationally constrained heterogeneous edge devices, FL must cope with uneven compute, memory, and communication capabilities; architectural adaptations, system-level scheduling, synchronous/asynchronous aggregation, distillation, and split learning are surveyed in \cite{pfeiffer2023federated}. Classical mitigations include FedProx \cite{li2020}, resource-aware adaptive FL \cite{wang2020}, FedCS \cite{nishio2019}, utility-based selection in Oort \cite{oort2021}, and tiered scheduling in TiFL \cite{tifl2020}. Energy- and hardware-oriented extensions further consider CPU--GPU co-resource management, asynchronous training, and differential-privacy-aware load modeling \cite{zeng2020energy,xu2021asynchronous,baligodugula2025hardwareaware}. Most of these methods do not jointly optimize client selection and local workload adaptation within the same framework; the proposed approach addresses both dimensions together.

For efficient training under client heterogeneity, FedEff assigns per-client local epoch budgets from compute and communication profiles via deterministic server-side rules, reducing wall-clock time and waiting delays \cite{narmadha2025fedeff}. Complementary approaches instead learn heterogeneous local training intensity policies with deep reinforcement learning, adapting per-round workload to client compute capacity and communication delay \cite{zeng2023hetintensity}. Our HW-FL complements such epoch adaptation with hardware-score-based top-$K$ selection and communication-aware aggregation in a single loop.

In affective computing, federated multimodal emotion recognition on edge devices has been explored with lightweight CNN-based audiovisual models, reporting improved classification under privacy-preserving training \cite{simic2024flser}. Centralized bimodal audio--text systems on IEMOCAP with data augmentation can also achieve high accuracy \cite{bimer2025}. Nevertheless, hardware-aware FL integration for speech emotion recognition in heterogeneous edge environments remains limited.

\section{System Model}

The system model considered in this study consists of a federated learning server and multiple connected edge clients. Clients store local data for the speech emotion recognition task without leaving the device and transmit only model parameter updates to the server. Client devices are heterogeneous in terms of CPU core count, memory capacity, and network latency, and these differences directly affect training efficiency.

In each federated learning round, the server broadcasts the current global model to selected clients; clients perform local training and send model updates back to the server. The server aggregates these updates to produce a new global model. The proposed hardware-aware mechanisms are integrated into the client selection and local training steps of this basic FL loop.

\section{Hardware-Aware Federated Learning Framework}

\begin{figure}[!t]
\centering
\resizebox{\columnwidth}{!}{\includegraphics{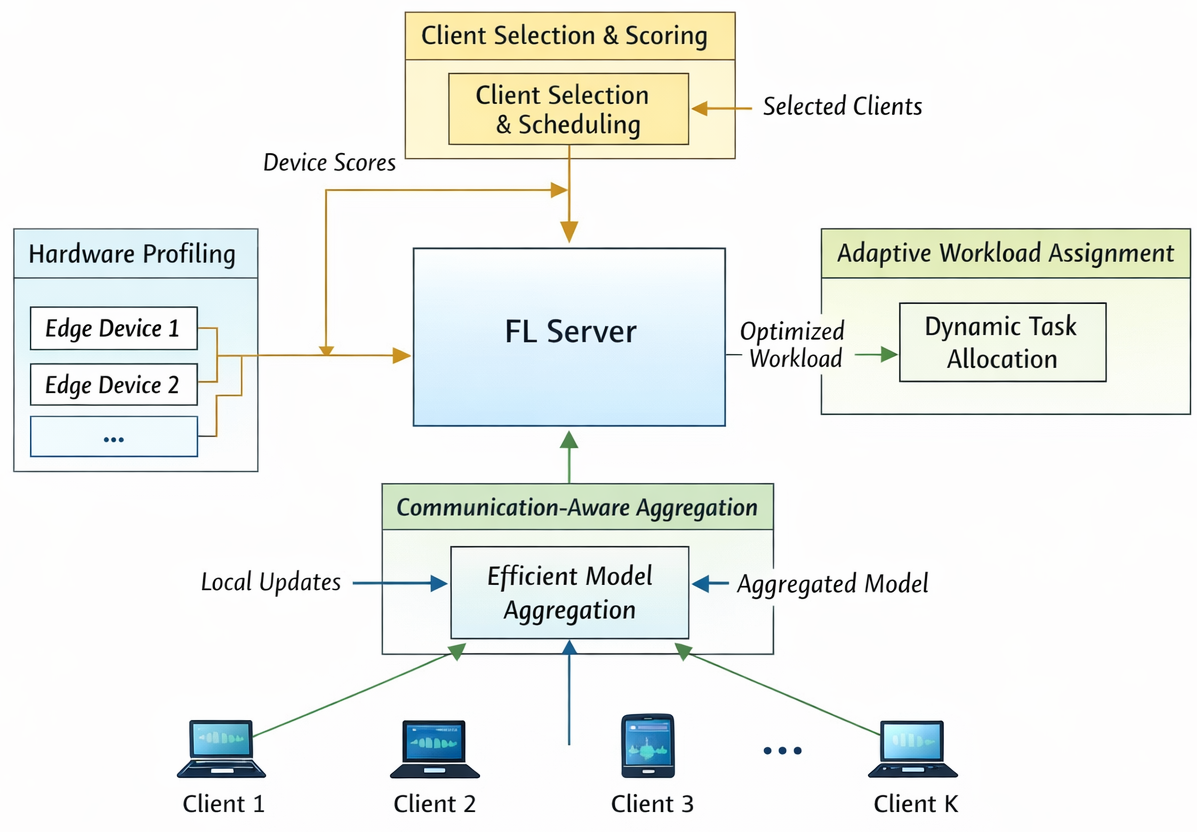}}
\caption{Overall workflow and components of the hardware-aware federated learning system.}
\label{fig:mimari}
\end{figure}

The hardware-aware federated learning framework proposed in this study addresses hardware profiling, client selection, and adaptive workload assignment within an integrated structure. The overall workflow and interactions among components are presented in Fig.~\ref{fig:mimari}.

\subsection{System Architecture}

The proposed system consists of multiple edge devices connected to a federated learning server. Each client performs local model training using private speech data and periodically sends model updates to the server. The server aggregates received updates to produce a global model. Edge devices include heterogeneous hardware platforms such as laptops, tablets, and smartphones.

\subsection{Hardware Profiling}

Each client device performs local hardware profiling to estimate computational capabilities. The hardware profile is represented as
\begin{equation}
H_i = (CPU_i, RAM_i, T_i, L_i)
\end{equation}
where $CPU_i$ denotes the number of CPU cores, $RAM_i$ available memory, $T_i$ model training latency, and $L_i$ network communication latency. To reduce the effect of scale differences, these quantities are normalized before scoring; for example,
\begin{equation}
\hat{CPU}_i = \frac{CPU_i}{\max_j CPU_j}, \quad \hat{L}_i = \frac{L_i}{\max_j L_j}
\end{equation}
can be defined.

\subsection{Hardware-Aware Client Planning and Optimization}

Based on collected hardware profiles, the federated server computes a hardware score for each client:
\begin{equation}
S_i = \alpha \cdot CPU_i + \beta \cdot RAM_i + \gamma \cdot \frac{1}{T_i} - \delta \cdot L_i
\end{equation}
where $\alpha$, $\beta$, $\gamma$, and $\delta$ are weights controlling the importance of processor, memory, training latency, and communication latency, respectively. This score can be interpreted as a proxy that rewards computational capacity and training efficiency while penalizing latency. Clients with higher scores are prioritized for participation in training. Federated round scheduling can be approximated by the following multi-objective problem:
\begin{equation}
\min_{S_t,\{E_i\}} \max_{i \in S_t} T_i(E_i) + \lambda \, C(S_t)
\end{equation}
Here, $S_t$ denotes the selected client set, $E_i$ the number of local epochs for client $i$, $T_i(E_i)$ local training time, $C(S_t)$ communication load, and $\lambda$ the trade-off coefficient. This formulation approximately expresses a constrained optimization problem that minimizes round completion time under communication and resource constraints. The problem has an NP-hard structure because client selection and workload assignment must be optimized jointly. Therefore, the proposed method uses a hardware-aware score-based heuristic to obtain a practical solution without increasing computational cost excessively. The heuristic showed stable learning behavior in experiments. However, theoretical convergence guarantees are left for future work.

\subsection{Adaptive Local Training}

To further improve training efficiency, the number of local training epochs is adjusted dynamically according to each client's hardware score:
\begin{equation}
E_i = E_{\text{base}} \times \frac{S_i}{S_{\max}}
\end{equation}
This mechanism enables stronger devices to contribute more effectively during federated training. The proposed framework is model-agnostic and can be applied to CNN, RNN, and transformer-based architectures with the same scheduling logic. In this study, a lightweight model is preferred primarily to isolate system-level contributions independent of model architecture.
To monitor long-term client inclusiveness, participation fairness can also be defined. Let $f_i$ be the selection frequency of client $i$; selection fairness can be measured by the Jain index
\begin{equation}
J = \frac{(\sum_i f_i)^2}{N \sum_i f_i^2}
\end{equation}
where values of $J$ closer to 1 indicate more balanced participation.

Alg.~\ref{alg:hwfl} summarizes the end-to-end HW-FL training loop that integrates hardware profiling, top-$K$ client selection, adaptive local epochs, and communication-aware aggregation.

\begin{algorithm}[!t]
\caption{Hardware-Aware Federated Learning (HW-FL)}
\label{alg:hwfl}
\footnotesize
\begin{algorithmic}[1]
\Require Initial global model $w^{(0)}$; $N$ clients with local data $D_i$; rounds $T$; top-$K$; $E_{\text{base}}$; weights $(\alpha,\beta,\gamma,\delta)$
\State \textbf{Profiling:} for each client $i$, obtain $H_i=(CPU_i,RAM_i,T_i,L_i)$ and compute $S_i$ (Eq.~3); let $S_{\max}=\max_i S_i$
\For{$t=1$ to $T$}
  \State Select $\mathcal{S}_t \leftarrow$ top-$K$ clients with highest $S_i$
  \State Broadcast $w^{(t-1)}$ to all $i \in \mathcal{S}_t$
  \ForAll{$i \in \mathcal{S}_t$ \textbf{in parallel}}
    \State $E_i \leftarrow E_{\text{base}} \cdot S_i / S_{\max}$ \Comment{Eq.~4}
    \State $w_i^{(t)} \leftarrow \textsc{LocalTrain}\bigl(w^{(t-1)}, D_i, E_i\bigr)$
  \EndFor
  \State $w^{(t)} \leftarrow \sum_{i \in \mathcal{S}_t} \dfrac{n_i S_i}{\sum_{j \in \mathcal{S}_t} n_j S_j}\, w_i^{(t)}$
\EndFor
\State \Return $w^{(T)}$
\end{algorithmic}
\end{algorithm}

\subsection{Communication and Complexity Analysis}

The main advantage of the proposed method appears not only in accuracy but also in system cost. The total communication cost per round can be approximated as
\begin{equation}
\mathcal{C}_r = (|S_r| + 1)\cdot |w|
\end{equation}
where $|S_r|$ is the number of selected clients and $|w|$ is the size of the model update. FedAvg uses all clients ($|S_r|=N$), whereas hardware-aware selection uses $|S_r|=K<N$, directly reducing communication cost. Similarly, in synchronous FL scenarios, round duration is approximately determined by the slowest selected client; systematically excluding high-latency clients reduces the \textit{straggler} effect. Since no real energy measurements were performed, energy consumption can be defined only as a proxy:
\begin{equation}
E_i^{\text{energy}} \propto CPU_i \times T_i
\end{equation}
Therefore, the proposed selection mechanism also has the potential to indirectly reduce energy consumption by prioritizing efficient clients; however, real energy measurement is outside the scope of this study.

\FloatBarrier
\section{Experimental Setup}

The proposed framework was evaluated on a unified CSV release of the IEMOCAP dataset \cite{iemocap}. Four basic emotion classes (ang, hap, sad, neu) were used. For the federated scenario, $N{=}5$ clients were created according to Session1--Session5 partitioning; this number follows the natural session structure of the dataset and provides a reproducible edge scenario while preserving a non-IID split.

The proposed data pipeline extracts 40-dimensional MFCC features from \texttt{wav\_path} (mean pooling along the time axis). When audio files are unavailable, the pipeline falls back to text-based feature extraction; audio was used for approximately 71.3\% of samples. The primary classifier is a lightweight 1D CNN, chosen to isolate system-level contributions. MFCC+$\Delta$ and lightweight BiLSTM variants are supported in the infrastructure and will be reported in future work for accuracy gains.

Three baselines and three ablation variants were compared. Oort, FedCS, and TiFL were discussed qualitatively. The proposed method uses top-$K{=}3$ selection per round, adaptive epochs, and communication-aware aggregation. All methods were evaluated over 50 rounds and 5 trials; communication cost was computed from the transmitted model size (MB). Representative device profiles in the simulation are given in Table~\ref{tab:cihaz}; latency values were derived from representative laptop/phone measurement ranges and additionally sampled from a log-normal distribution per round in supplementary experiments.

\begin{table}[!t]
\centering
\caption{\textsc{Representative Edge Device Profiles}\label{tab:cihaz}}
\resizebox{\columnwidth}{!}{%
\begin{tabular}{|l|c|c|c|}
\hline
Device & CPU & RAM (GB) & Latency (ms) \\
\hline
Laptop (high-end) & 16 & 32 & 170 \\
Tablet & 4 & 32 & 183 \\
Phone (low RAM) & 4 & 16 & 200 \\
Legacy phone & 2 & 32 & 261 \\
Phone (low latency) & 4 & 32 & 132 \\
\hline
\end{tabular}%
}
\end{table}

Hardware scoring used \(\alpha{=}0{,}4\), \(\beta{=}0{,}2\), \(\gamma{=}0{,}3\), \(\delta{=}0{,}1\): \(\alpha\) and \(\gamma\) emphasize CPU and training efficiency to limit stragglers; \(\beta\) addresses memory constraints; \(\delta\) penalizes network latency. In a sensitivity sweep over \(\alpha\in\{0.3,0.4,0.5\}\), mean accuracy differed by less than $0.02$. In participant scaling experiments with $K\in\{1,\ldots,5\}$, accuracy did not increase monotonically; at $K{=}1$ it reached 0.40, whereas for $K{\geq}2$ it remained in the $\sim$0.32--0.35 band (Table~\ref{tab:scale}).

\begin{table}[!t]
\centering
\caption{\textsc{HW-FL Accuracy vs.\ Participant $K$ (5 Trials)}\label{tab:scale}}
\resizebox{\columnwidth}{!}{%
{\fontsize{8}{9}\selectfont
\begin{tabular}{|c|c|c|c|c|c|}
\hline
$K$ & 1 & 2 & 3 & 4 & 5 \\
\hline
Acc. & 0.40 & 0.35 & 0.32 & 0.34 & 0.34 \\
\hline
\end{tabular}}%
}
\end{table}

Experiments were repeated with 5 different random seeds, and speech emotion recognition performance was evaluated using accuracy, macro-F1, and balanced accuracy. The system was developed in Python and PyTorch. The client count was limited to $N{=}5$ because of the number of sessions in IEMOCAP. Validation on larger-scale scenarios and real devices (Windows, Android, and iOS) will be addressed in future work.

\section{Results and Analysis}

Experimental results show that the proposed approach is more efficient in heterogeneous client environments. For the SER task, HW-FL achieved accuracy $0.352{\pm}0.019$, macro-F1 $0.252{\pm}0.060$, and balanced accuracy $0.305{\pm}0.031$; FedAvg obtained $0.325{\pm}0.032$, $0.262{\pm}0.048$, and $0.314{\pm}0.030$, respectively. The random-guess baseline in the four-class non-IID setup is 0.25.

\begin{table}[!t]
\centering
\caption{\textsc{Baseline Comparison (50 Rounds, 5 Trials)}\label{tab:performans}}
\resizebox{\columnwidth}{!}{%
\begin{tabular}{|l|c|c|c|}
\hline
Method & Acc. (mean$\pm$std) & Time (s) & Comm. (MB) \\
\hline
FedAvg & 0.325$\pm$0.032 & 12.11$\pm$8.18 & 23.81 \\
FedProx & 0.325$\pm$0.032 & 7.37$\pm$0.29 & 23.81 \\
Random top-$K$ & 0.348$\pm$0.009 & 7.42$\pm$0.29 & 14.29 \\
HW-FL (proposed) & 0.352$\pm$0.019 & 7.69$\pm$0.24 & 14.29 \\
\hline
\end{tabular}
}
\end{table}

\begin{figure}[!t]
\centering
\resizebox{\columnwidth}{!}{\includegraphics{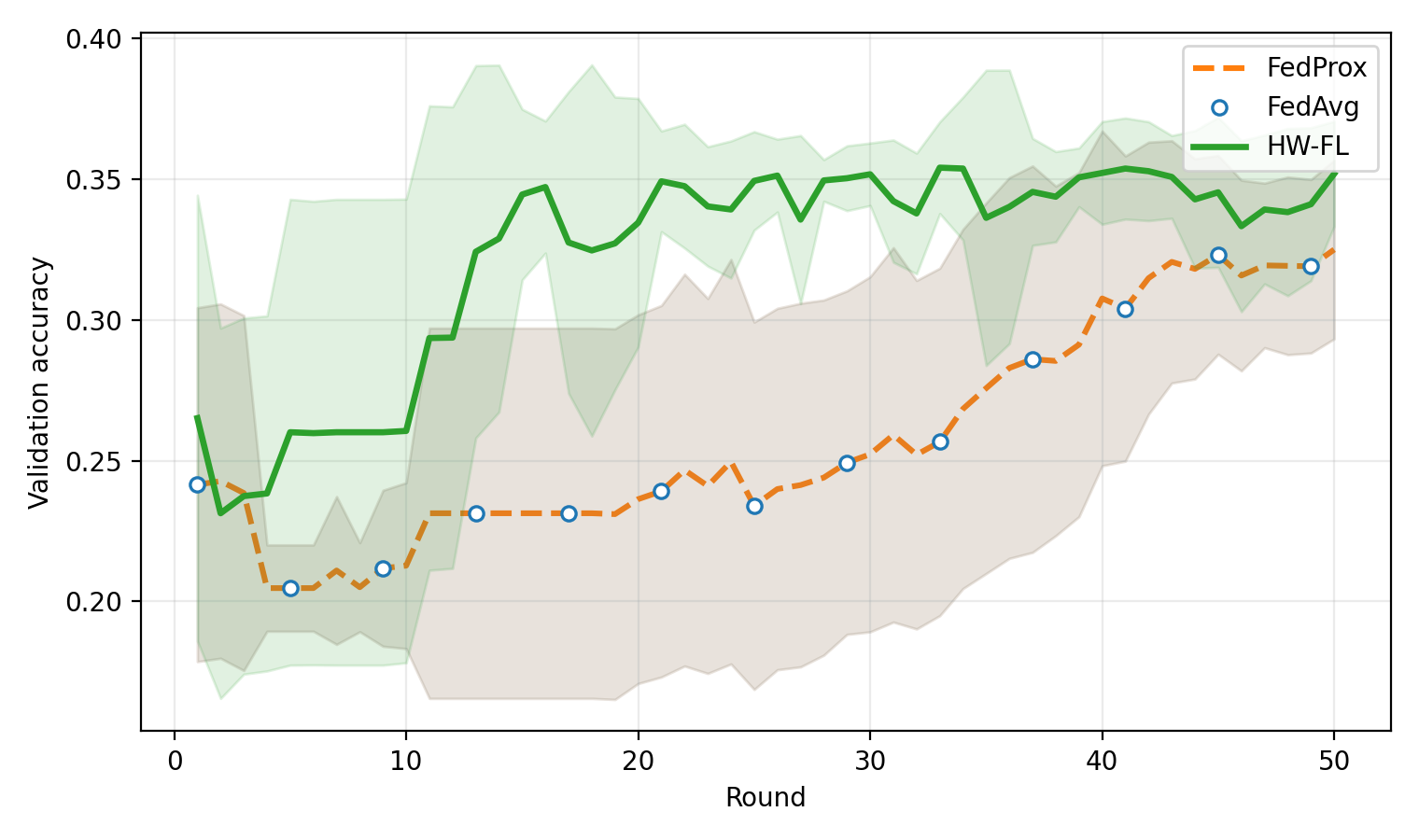}}
\caption{Round-wise accuracy curves for major methods.}
\label{fig:workflow}
\end{figure}

\begin{figure}[!t]
\centering
\resizebox{\columnwidth}{!}{\includegraphics{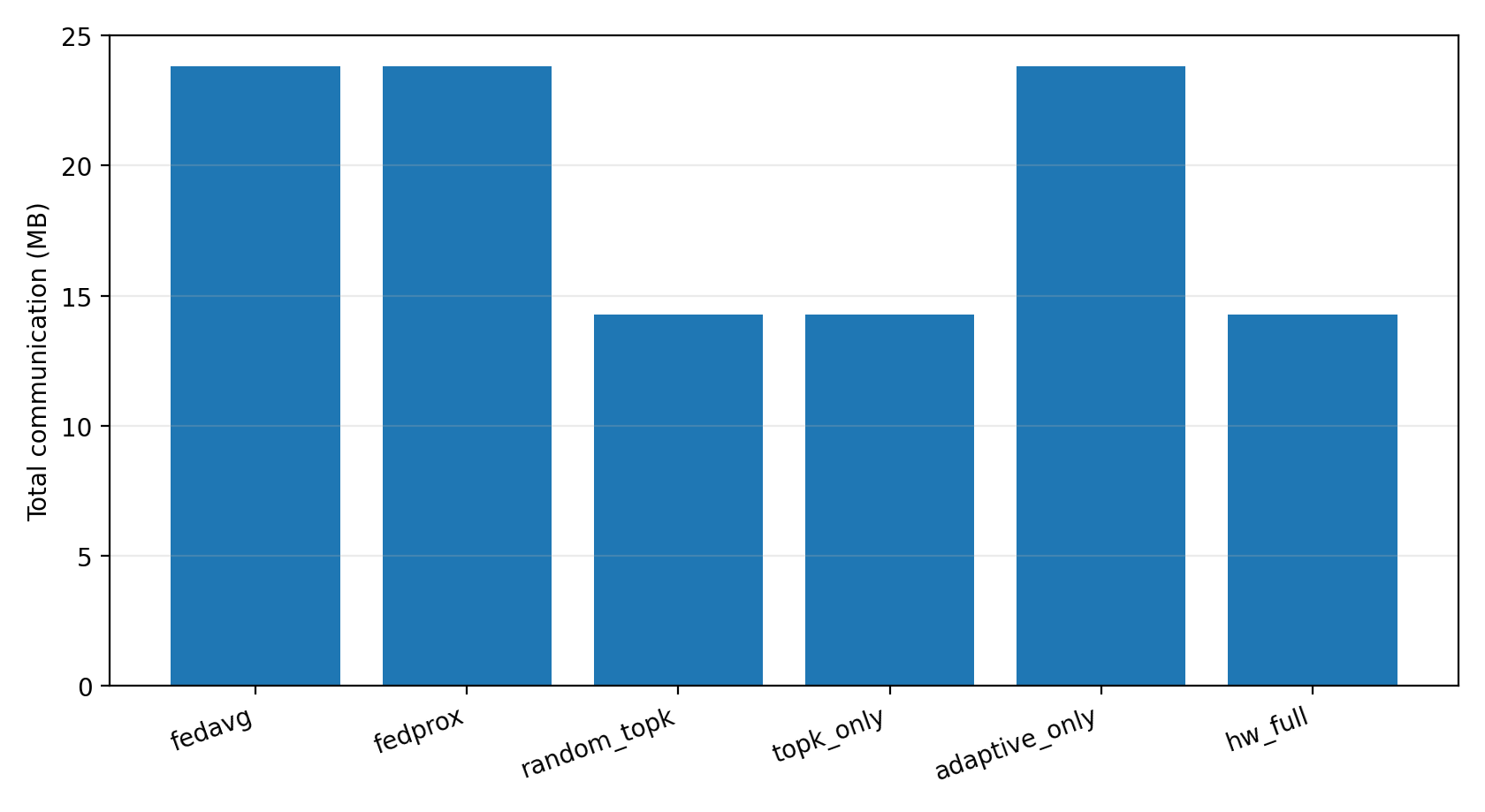}}
\caption{Total communication cost comparison across methods.}
\label{fig:comm}
\end{figure}

Table~\ref{tab:performans} shows total time decreasing from $12.11{\pm}8.18$\,s to $7.69{\pm}0.24$\,s ($\sim$36.5\% reduction) and communication from 23.81\,MB to 14.29\,MB. Convergence is shown in Fig.~\ref{fig:workflow} and communication in Fig.~\ref{fig:comm}.

Random top-$K$ showed accuracy close to HW-FL with lower variance. In a Welch $t$-test, the HW-FL--FedAvg accuracy difference was not significant at the 5\% level ($p{=}0.19$); however, the effect size was moderate ($d{\approx}0.92$). Gains in system metrics (time/communication) were more consistent.

\section{Ablation Study}

This study separately examines two core components of the full hardware-aware framework: (i) top-$K$ client selection based only on hardware score, and (ii) adaptive epochs only. Results show that each component alone achieves better or comparable accuracy than FedAvg with lower runtime.

\begin{table}[!t]
\centering
\caption{\textsc{Ablation Results (mean$\pm$std)}\label{tab:ablation}}
\resizebox{\columnwidth}{!}{%
\begin{tabular}{|l|c|c|c|}
\hline
Variant & Acc. & Time (s) & Comm. (MB) \\
\hline
FedAvg & 0.325$\pm$0.032 & 12.11$\pm$8.18 & 23.81 \\
Top-$K$ only & 0.284$\pm$0.045 & 7.81$\pm$0.55 & 14.29 \\
Adaptive-only & 0.358$\pm$0.010 & 8.19$\pm$0.41 & 23.81 \\
Full HW-FL & 0.352$\pm$0.019 & 7.69$\pm$0.24 & 14.29 \\
\hline
\end{tabular}
}
\end{table}

Table~\ref{tab:ablation} shows the effect of each component separately. When only top-$K$ client selection is used, accuracy (0.284) remains significantly lower than full HW-FL (0.352) ($p{<}0.05$). This indicates that client selection alone is insufficient.

The adaptive-epoch-only configuration reaches the highest accuracy (0.358) but increases communication cost. In contrast, full HW-FL offers a more balanced trade-off among accuracy, training time, and communication cost.

When round durations are examined, FedAvg and FedProx include all clients in every round, leading to higher variance; methods with hardware-aware selection exhibit narrower distributions. This supports the claim that the straggler effect is reduced.

\section{Discussion and Conclusion}

\begin{spacing}{1.03}
The results show that hardware-aware scheduling can improve federated learning efficiency in heterogeneous edge device environments. Top-$K$ client selection reduced the number of participants while improving both SER performance and system metrics. Although absolute accuracy ($\sim$0.35) exceeds the random baseline (0.25) under a four-class non-IID FL setup with a lightweight CNN, the main contribution of this work is optimizing time and communication rather than accuracy alone.
The main limitations are the $N{=}5$ session split, simulation-based hardware profiles, and the lack of real-device validation. Future work will explore MFCC+$\Delta$/BiLSTM-based models, scalability analyses with $N{\gg}5$, and real-device tests on Windows, Android, and iOS platforms.
\end{spacing}
\vspace{-0.1em}
{\tiny
\setlength{\itemsep}{0pt}
\setlength{\parskip}{0pt}

}


\begin{thebibliography}{99}

\bibitem{mcmahan2017}
B. McMahan, E. Moore, D. Ramage, S. Hampson, B. A. y Arcas, ``Communication-efficient learning of deep networks from decentralized data,'' \textit{Proc. AISTATS}, 2017.
\bibitem{kairouz2021}
P. Kairouz et al., ``Advances and open problems in federated learning,'' \textit{Found. Trends Mach. Learn.}, 2021.
\bibitem{li2020}
T. Li, A. K. Sahu, M. Zaheer, M. Sanjabi, A. Talwalkar, V. Smith, ``Federated optimization in heterogeneous networks,'' \textit{Proc. MLSys}, 2020.
\bibitem{wang2020}
S. Wang, T. Tuor, T. Salonidis et al., ``Adaptive federated learning in resource-constrained edge computing systems,'' \textit{IEEE J. Sel. Areas Commun.}, 2020.
\bibitem{nishio2019}
T. Nishio and R. Yonetani, ``Client selection for federated learning with heterogeneous resources,'' \textit{Proc. IEEE ICC}, 2019.
\bibitem{iemocap}
C. Busso et al., ``IEMOCAP: Interactive emotional dyadic motion capture database,'' \textit{Lang. Resour. Eval.}, vol. 42, no. 4, pp. 335--359, 2008.
\bibitem{bimer2025}
E. Dikbiyik, O. Demir, and B. Dogan, ``BiMER: Design and implementation of a bimodal emotion recognition system enhanced by data augmentation techniques,'' \textit{IEEE Access}, vol. 13, pp. 64330--64352, 2025, doi: 10.1109/ACCESS.2025.3559339.
\bibitem{bonawitz2019}
K. Bonawitz et al., ``Towards federated learning at scale: System design,'' \textit{Proc. MLSys}, 2019.
\bibitem{oort2021}
F. Lai et al., ``Oort: Efficient federated learning via guided participant selection,'' \textit{Proc. OSDI}, 2021.
\bibitem{tifl2020}
Z. Chai et al., ``TiFL: A tier-based federated learning system,'' \textit{Proc. HPDC}, 2020.
\bibitem{zeng2020energy}
Q. Zeng et al., ``Energy-efficient resource management for federated edge learning with CPU-GPU heterogeneous computing,'' \textit{IEEE Trans. Wireless Commun.}, vol. 20, pp. 7947--7962, 2020.
\bibitem{xu2021asynchronous}
C. Xu et al., ``Asynchronous federated learning on heterogeneous devices: A survey,'' \textit{Comput. Sci. Rev.}, vol. 50, pp. 100595, 2021.
\bibitem{pfeiffer2023federated}
K. Pfeiffer et al., ``Federated learning for computationally constrained heterogeneous devices: A survey,'' \textit{ACM Comput. Surv.}, vol. 55, pp. 1--27, 2023, doi: 10.1145/3596907.
\bibitem{simic2024flser}
N. Simi\'{c} et al., ``Enhancing emotion recognition through federated learning: A multimodal approach with convolutional neural networks,'' \textit{Appl. Sci.}, vol. 14, no. 4, 2024, doi: 10.3390/app14041325.
\bibitem{narmadha2025fedeff}
K. Narmadha and P. Varalakshmi, ``FedEff: Efficient federated learning with optimal local epochs for heterogeneous clients,'' \textit{Sci. Rep.}, vol. 15, 2025, doi: 10.1038/s41598-025-22672-1.
\bibitem{zeng2023hetintensity}
M. Zeng, X. Wang, W. Pan, and P. Zhou, ``Heterogeneous training intensity for federated learning: A deep reinforcement learning approach,'' \textit{IEEE Trans. Netw. Sci. Eng.}, vol. 10, no. 2, pp. 990--1002, 2023.
\bibitem{baligodugula2025hardwareaware}
V. Baligodugula and F. H. Amsaad, ``Hardware-aware federated learning: Optimizing differential privacy in distributed computing architectures,'' \textit{Electronics}, 2025, doi: 10.3390/electronics14061218.
\bibitem{YUKSEL2026109557}
B. B. Yuksel and A. Y. Metin, ``Federated learning with homomorphic encryption for secure real time ECG anomaly detection: A multi institutional privacy preserving framework,'' \textit{Biomed. Signal Process. Control}, vol. 116, pp. 109557, 2026.

\end{thebibliography}
\end{document}